\documentclass[sigconf, balance=True]{acmart}

\usepackage{enumitem}
\usepackage{makecell}
\usepackage{tabulary}
\usepackage[table]{xcolor}
\usepackage{colortbl}
\usepackage{multirow}
\usepackage{tabularx}
\usepackage{array}
\usepackage{graphicx}
\newcolumntype{Y}{>{\centering\arraybackslash\hspace{0pt}\nohyphens}X}
\usepackage{stfloats}
\usepackage{balance}
\definecolor{shallowblue}{RGB}{227, 240, 249}
\definecolor{deepblue}{RGB}{192, 216, 240}
\AtBeginDocument{%
  }

\copyrightyear{2026}
\acmYear{2026}
\setcopyright{cc}
\setcctype{by}
\acmConference[MM '26]{Proceedings of the 34th ACM International Conference on Multimedia}{November 10--14, 2026}{Rio de Janeiro, Brazil}
\acmBooktitle{Proceedings of the 34th ACM International Conference on Multimedia (MM '26), November 10--14, 2026, Rio de Janeiro, Brazil}
\acmDOI{10.1145/3767308.3835022}
\acmISBN{979-8-4007-2213-4/2026/11}
\begin{document}

\title{Bridging Time and Space: Decoupled Spatio-Temporal Alignment for Video Grounding}

\settopmatter{authorsperrow=4}
\author{Xuezhen Tu}
\affiliation{%
  \institution{\mbox{Shanghai Jiao Tong University}}
  \city{Shanghai}
  \country{China}
}
\email{xuezhentu@sjtu.edu.cn}

\author{Jingyu Wu}
\affiliation{%
  \institution{ZTE Corporation}
  \city{Nanjing}
  \country{China}}
\email{wu.jingyu2@zte.com.cn}

\author{Fangyu Kang}
\affiliation{%
  \institution{ZTE Corporation}
  \city{Nanjing}
  \country{China}}
\email{kang.fangyu@zte.com.cn}

\author{Qingpeng Nong}
\affiliation{%
  \institution{ZTE Corporation}
  \city{Nanjing}
  \country{China}}
\email{nong.qingpeng@zte.com.cn}

\author{Kaijin Zhang}
\affiliation{%
  \institution{ZTE Corporation}
  \city{Nanjing}
  \country{China}}
\email{zhang.kaijin1@zte.com.cn}

\author{Chaoyue Niu}
\authornote{Chaoyue Niu and Fan Wu are corresponding authors.}
\correspondingauthor
\affiliation{
  \institution{
    \hspace*{-1.4em}
    \mbox{Shanghai Jiao Tong University}
  }
  \city{Shanghai}
  \country{China}
}
\email{rvince@sjtu.edu.cn}

\author{
  \texorpdfstring{
    \advance\leftskip by 1.2em\relax
    Fan Wu}{Fan Wu}
}
\authornotemark[1]
\correspondingauthor
\affiliation{%
  \institution{\mbox{Shanghai Jiao Tong University}}
  \city{Shanghai}
  \country{China}
}
\email{fwu@cs.sjtu.edu.cn}



\renewcommand{\shortauthors}{Xuezhen Tu et al.}

\begin{abstract}
Spatio-Temporal Video Grounding requires jointly localizing target objects across both temporal and spatial dimensions based on natural language queries, posing fundamental challenges for existing Multimodal Large Language Models (MLLMs). We identify two core challenges: \textit{entangled spatio-temporal alignment}, arising from coupling two heterogeneous sub-tasks within the same autoregressive output space, and \textit{dual-domain visual token redundancy}, where target objects exhibit simultaneous temporal and spatial sparsity, rendering the overwhelming majority of visual tokens irrelevant to the grounding query.
To address these, we propose \textbf{Bridge-STG}, an end-to-end framework that decouples temporal and spatial localization while maintaining semantic coherence. While decoupling is the natural solution to this entanglement, it risks creating a semantic gap between the temporal MLLM and the spatial decoder. Bridge-STG resolves this through two pivotal designs: the \textbf{Spatio-Temporal Semantic Bridging (STSB)} mechanism with Explicit Temporal Alignment (ETA) distills the MLLM's temporal reasoning context into enriched bridging queries as a robust semantic interface; and the \textbf{Query-Guided Spatial Localization (QGSL)} module leverages these queries to drive a purpose-built spatial decoder with multi-layer interactive queries and positive/negative frame sampling, jointly eliminating dual-domain visual token redundancy.
Extensive experiments across multiple benchmarks demonstrate that Bridge-STG achieves state-of-the-art performance among MLLM-based methods. Bridge-STG improves average m\_vIoU from $26.4$ to $34.3$ on VidSTG and demonstrates strong cross-task transfer across various fine-grained video understanding tasks under a unified multi-task training regime.

\end{abstract}



\begin{teaserfigure}
  \includegraphics[width=\textwidth]{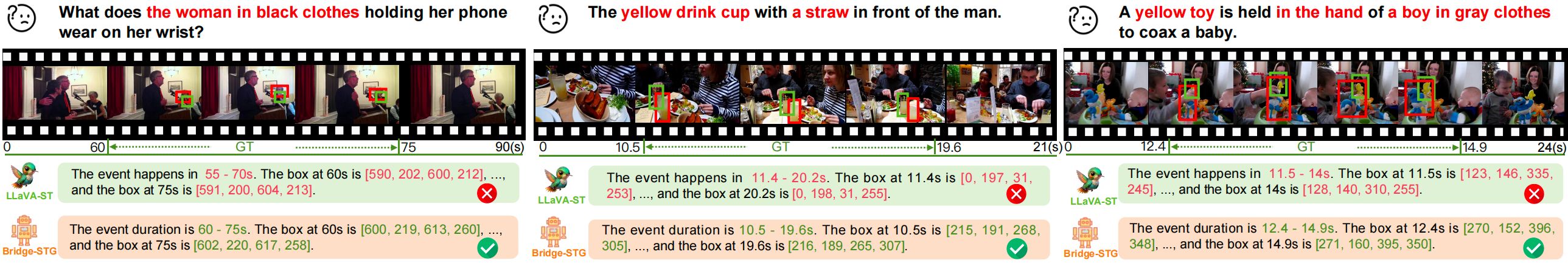}
  \caption{Comparison of MLLM-based models on the Spatio-Temporal Video Grounding (STVG) task. Current MLLM-based methods struggle with complex scenarios, specifically through inaccurate temporal grounding, confusion with similar distractors, and spatial-semantic misalignment. Content in \textcolor{green}{green} and \textcolor{red}{red} represents correct and wrong answers, respectively.}
  \label{fig:teaser}
\end{teaserfigure}



\maketitle

\section{Introduction}

Spatio-Temporal Video Grounding (STVG) aims to identify and locate target objects across both the temporal and spatial dimensions of a video based on natural language queries~\cite{zhang2020does}. 
It has widespread utility in fields such as autonomous driving~\cite{ahmad2025videomolmo,vishal2024eyes}, video retrieval~\cite{gu2024context,yang2025multi}, and intelligent surveillance~\cite{zhang2020does}. 
Recently, Multimodal Large Language Models (MLLMs) significantly propel this task forward owing to their superior capabilities in multimodal semantic comprehension and structured reasoning~\cite{comanici2025gemini,grattafiori2024llama,wang2025capabilities,yang2025qwen3}.

Despite these advances, there is still a significant gap between current MLLM-based methods and real-world demands~\cite{ahmad2025videomolmo,li2025llava,wang2025spacevllm}. 
As illustrated in Fig.~\ref{fig:teaser}, existing MLLMs frequently fail in complex scenarios: producing temporally imprecise boundaries, confusing the target object with visually similar distractors, and generating outputs that are semantically plausible yet spatially mislocalized. 
While recent works attempt to narrow this gap, they predominantly rely on MLLM-generated query embeddings for spatial grounding within coupled architectures.
The lack of temporal-spatial decoupling bottlenecks their fine-grained spatial localization precision.

We attribute these limitations to two challenges. 
The first is \textbf{entangled spatio-temporal alignment}. STVG inherently comprises two heterogeneous sub-tasks: temporal localization, which requires high-level semantic reasoning over event sequences and is naturally suited to MLLMs~\cite{maaz2024video,ren2024timechat}, and spatial localization, which demands pixel-precise coordinate prediction and is better served by specialized detection architectures~\cite{zhu2020deformable,liu2024grounding}. 
Existing MLLMs that couple both tasks within the same autoregressive output space fail to exploit LLMs' temporal reasoning strengths, while their coordinate regression objective, trained with cross entropy over discretized tokens, is not suited for continuous spatial precision. Furthermore, the exponentially expanded joint spatio-temporal output space, compounded by complex temporal dynamics such as variable event durations, asynchronous activities, and scene transitions, further destabilizes multimodal alignment~\cite{yang2022tubedetr, su2021stvgbert}.

The second challenge is \textbf{dual-domain visual token redundancy}. Unlike image-based grounding, STVG suffers from redundancy in both temporal and spatial dimensions simultaneously. 
Target objects are present only within a fraction of the video duration (temporal sparsity), and even within that window they occupy only a localized spatial region (spatial sparsity)~\cite{jin2024chat,yang2025pvc}. 
This dual sparsity means that the overwhelming majority of visual tokens extracted from densely sampled frames are doubly irrelevant to the grounding query, severely obscuring the fine-grained correspondence between language descriptions and target locations and substantially degrading spatial localization fidelity~\cite{ma2024groma,zhang2024llava}.

While decoupled architecture is the natural solution to break the entangled alignment, it inherently creates a semantic gap between the temporal MLLM and the spatial decoder~\cite{lai2024lisa,ahmad2025videomolmo}.
To address these challenges, we propose {\textbf{Bridge-STG (Bridge-based Spatio-Temporal Video Grounding Model)}}, an end-to-end framework that decouples temporal and spatial localization while ensuring semantic coherence.
Bridge-STG is driven by two pivotal designs:

First, to bridge the architectural semantic gap, we introduce the \textbf{Spatio-Temporal Semantic Bridging (STSB)} mechanism. 
This process begins with an Explicit Temporal Alignment (ETA) strategy, which injects text-formatted timestamp tokens as virtual spatial coordinates into the MLLM's embedding space. 
The ETA provides structured temporal anchoring, allowing the MLLM to establish a clear event-boundary perception without disrupting its continuous positional space.
Building upon this, STSB uses a set of learnable bridging queries that propagate through the MLLM's layers.
These queries distill the accumulated temporal reasoning context into semantically enriched features. By translating the MLLM's sequence-level understanding into spatio-temporal aware query embeddings, STSB enables cooperative optimization between the otherwise isolated temporal and spatial modules.

Second, to tackle the dual-domain visual token redundancy, we propose the \textbf{Query-Guided Spatial Localization (QGSL)} module. 
Rather than forcing the MLLM to autoregressively generate coordinate predictions, QGSL uses the semantic bridging queries from STSB as conditional prompts to drive a purpose-built spatial decoder.
To capture fine-grained multi-scale visual cues, QGSL incorporates multi-layer interactive queries that aggregate candidate features across all image encoder layers, enriching spatial feature diversity for localizing small or occluded objects.

Furthermore, QGSL is strengthened by a positive/negative frame sampling strategy during training. 
By intentionally exposing the spatial decoder to negative frames alongside the positive event frames, it forces the decoder to discriminate the target object from visually similar background distractors. 
This training mechanism and customized decoder design effectively filter out the overwhelming redundancy of irrelevant visual tokens, yielding precise instance-level spatial grounding.


In summary, our main contributions are as follows:
\begin{itemize}
\item We propose Bridge-STG, an end-to-end decoupled MLLM framework for fine-grained STVG. We introduce the STSB mechanism with Explicit Temporal Alignment (ETA) to distill temporally aligned reasoning context into robust bridging queries, effectively mitigating the semantic isolation of decoupled architectures.
\item We design a customized spatial decoding module, Query-Guided Spatial Localization (QGSL), which uses multi-layer interactive queries and a positive/negative frame sampling strategy. This design filters out dual-domain visual token redundancy, vastly improving spatial localization fidelity.
\item Extensive experiments demonstrate that Bridge-STG achieves state-of-the-art performance among MLLM-based methods on STVG task, improving the average m\_vIoU from 26.4 to 34.3. Furthermore, Bridge-STG exhibits strong cross-task transfer across diverse fine-grained video understanding tasks, matching or even surpassing task-specific models under a unified multi-task training regime.

\end{itemize}

\begin{figure*}
  \centering
  \includegraphics[width=\textwidth]{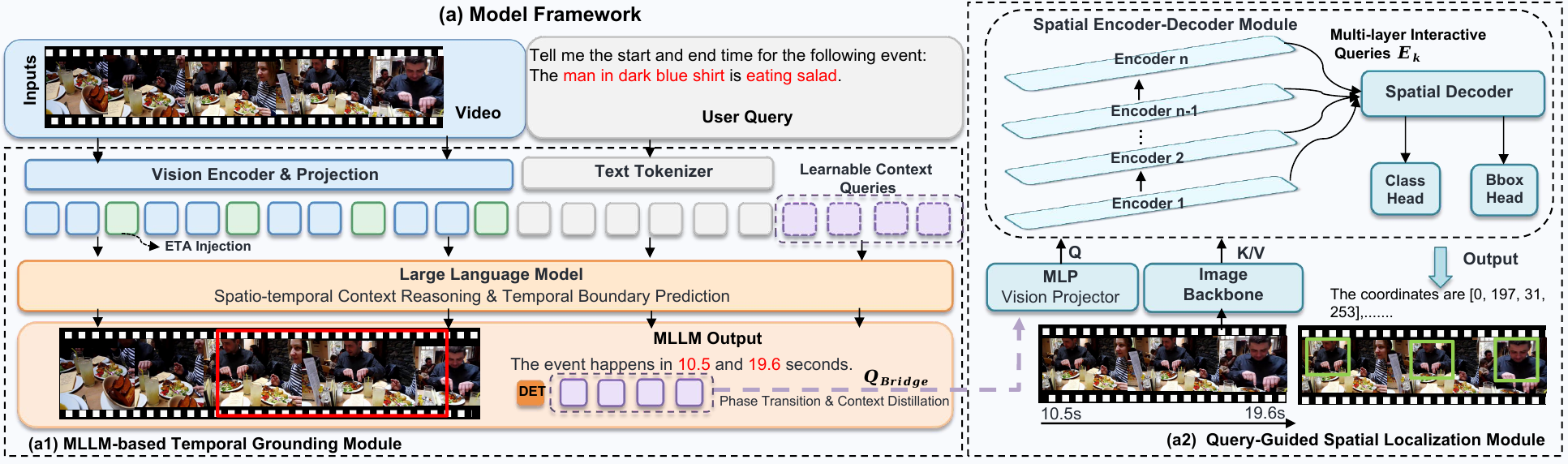}
  \caption{Overall architecture of Bridge-STG. The model first predicts the event's temporal window with the ETA strategy. Triggered by the [DET] token, the Spatio-Temporal Semantic Bridging mechanism distills the MLLM's reasoning context into bridging queries ($Q_{bridge}$). 
  Finally, the QGSL module utilizes $Q_{bridge}$ to perform precise spatial grounding on the located frames.
  }
  \label{fig:model_archi}
\end{figure*}

\section{Related Work}
\subsection{Spatio-Temporal Video Grounding}
 STVG~\cite{zhang2020does} aims to localize a target object in both space and time within a video based on a language query. The evolution of STVG methodologies can be categorized into two primary stages. Early frameworks~\cite{zhang2020does, zhang2020object, su2021stvgbert} adopt a sequential pipeline, where object proposals are initially generated by a pre-trained detector, followed by a selection mechanism to identify the correct one based on the linguistic query. In contrast, recent approaches~\cite{gu2024context, yang2022tubedetr, jin2022embracing, lin2023collaborative, yao2025omnistvg} have shifted toward an integrated encoder-decoder architecture, bypassing the dependency on external detection modules. Within this unified paradigm, the encoder is responsible for integrating multimodal cues from videos and text, and the decoder directly regresses the target's spatio-temporal coordinates, leading to enhanced performance. CG-STVG~\cite{gu2024context} and TubeDETR~\cite{yang2022tubedetr} employ zero-initialized object queries, which lack target-specific cues and thus struggle to learn discriminative target information from multimodal features in complex scenarios such as distractors or occlusion. In contrast, more recent TA-STVG~\cite{gu2025knowing} proposes a novel target-aware Transformer for STVG that adaptively generates object queries by exploring target-specific cues from the given video-text pair. 
 Recent extensions Video-GroundingDINO~\cite{wasim2024videogrounding} further explore open-vocabulary STVG by adapting detection transformers. 
 Despite these advances, existing methods still struggle with complex reasoning and open-vocabulary STVG due to their limited semantic capacity. In light of this, we explore MLLMs to inject stronger semantic understanding into the STVG task, since their extensive pretrained knowledge allows for better interpretation of complex language and adaptation to open-world scenarios.

\subsection{MLLMs for Grounding}
Recent advances in MLLMs~\cite{achiam2023gpt, ma2024groma, bai2025qwen3, guo2025seed1, guo2025deepseek, zeng2025glm, zhu2025internvl3} have yielded notable progress in visual grounding tasks. MiniGPT~\cite{chen2023minigpt}, LLaVA~\cite{zhang2024llava}, Qwen3-VL~\cite{bai2025qwen3}, and InternVL~\cite{zhu2025internvl3} concentrate on spatial grounding within static images, typically by generating bounding box coordinates or selecting from region proposals. For video grounding, certain works based on the aforementioned MLLM~\cite{guo2025vtg, guo2024trace, wang2023protege, barrios2023localizing, chen2021end} incorporate temporal grounding abilities, linking textual descriptions of events or actions to specific temporal spans through the prediction of start and end moments. VTG-LLM~\cite{guo2025vtg} introduces a slot-based token compression technique to enhance timestamp localization, while TRACE~\cite{guo2024trace} presents a causal event modeling framework designed to identify event timestamps. Recent works~\cite{ahmad2025videomolmo, li2025llava, wang2025spacevllm, gu2025thinking} explore MLLMs for joint spatio-temporal localization. VideoMolmo~\cite{ahmad2025videomolmo} initially predicts precise pointing coordinates by MLLM and employs a sequential mask-fusion module to generate coherent segmentation. LLaVA-ST~\cite{li2025llava} enables simultaneous spatio-temporal coordinate output via a progressive training strategy, yet its autoregressive paradigm yields ambiguous supervision, compromising spatial precision. SpaceVLLM~\cite{wang2025spacevllm} designs spatio-temporal-aware queries and a spatial decoder to help MLLM capture spatio-temporal information, while STVG-o1~\cite{gu2025thinking} employs a bounding-box chain-of-thought mechanism for explicit spatio-temporal reasoning. Concurrent work also explores zero-shot STVG~\cite{yang2025unleashing} and agentic reasoning frameworks~\cite{zhao2026agentic}. Despite these advances, they either suffer from the inherent spatial imprecision of coupled autoregressive generation~\cite{li2025llava} or lack an explicit semantic bridge between the MLLM's sequence-level temporal reasoning and the instance-level spatial decoding phase~\cite{ahmad2025videomolmo,wang2025spacevllm}. To address this, we propose Bridge-STG, which decouples the architecture and introduces the Spatio-Temporal Semantic Bridging mechanism to ensure semantic coherence across the architectural divide. Furthermore, we design the Query-Guided Spatial Localization module, integrating multi-layer interactive queries and P/N Frame Sampling to eliminate visual token redundancy and achieve precise spatial grounding.

\section{Method}
In this section, we describe the implementation details of Bridge-STG.
We first introduce the overall pipeline in Sec.~\ref{subsec:overview} and then describe the design of Explicit Temporal Alignment (ETA) module and Spatio-Temporal Semantic Bridging (STSB) in Sec.~\ref{subsec:dtsa}.
Subsequently, Sec.~\ref{subsec:qgsl} details the Query-Guided Spatial Localization (QGSL) module and Sec.~\ref{subsec:overall_loss} presents the overall training objectives.

\subsection{Overview}
\label{subsec:overview}
The overall architecture of Bridge-STG is shown in Fig.~\ref{fig:model_archi}. It is an end-to-end decoupled architecture for fine-grained STVG.
To overcome the inherent semantic gap caused by decoupling, Bridge-STG integrates a dedicated spatial localization module with temporal grounding via a robust semantic bridging mechanism.

Specifically, given an input video $\mathcal{V}$, frames are uniformly sampled at 2 fps, following standard practice in STVG~\cite{gu2024context,wang2025spacevllm}.
Every two consecutive frames are grouped into a frame pair. 
This paired grouping could capture short-term motion cues between adjacent seconds while halving the number of visual token groups, balancing performance with computational efficiency. 
Each frame pair is processed by the vision encoder and patch merger to produce a visual token representation, yielding the visual token sequence $\mathbf{V}_{feat} = \{\mathbf{v}_i\}_{i=1}^{N/2}$, where $N$ denotes the total number of sampled frames and $\mathbf{v}_i$ represents visual features for the $i$-th frame pair. 
Meanwhile, the user's text query $\mathcal{Q}$ is tokenized and encoded by the MLLM's text encoder to produce the corresponding text features $\mathbf{T}_{feat}$. 
The ETA module then injects text-formatted timestamps following each frame pair's visual tokens, providing the MLLM with structured temporal anchoring. 
MLLM subsequently processes $\mathbf{V}_{feat}$, the timestamp tokens, and $\mathbf{T}_{feat}$ jointly to predict the temporal window $[t_{start}, t_{end}]$ of the target event.

Upon completing temporal localization, the STSB is triggered by a special token \texttt{[DET]}.
Through a set of learnable bridging queries, the STSB learns the temporal reasoning context of the MLLM and transforms it into semantically enriched representations $\mathbf{Q}_{bridge}$, serving as the semantic interface to the downstream QGSL module.
Finally, QGSL takes $\mathbf{Q}_{bridge}$ as conditional prompts to perform precise spatial grounding on frames within $[t_{start}, t_{end}]$, producing the final per-frame bounding boxes.
Each component is detailed below.

\begin{figure*}
  \centering
  \includegraphics[width=\linewidth]{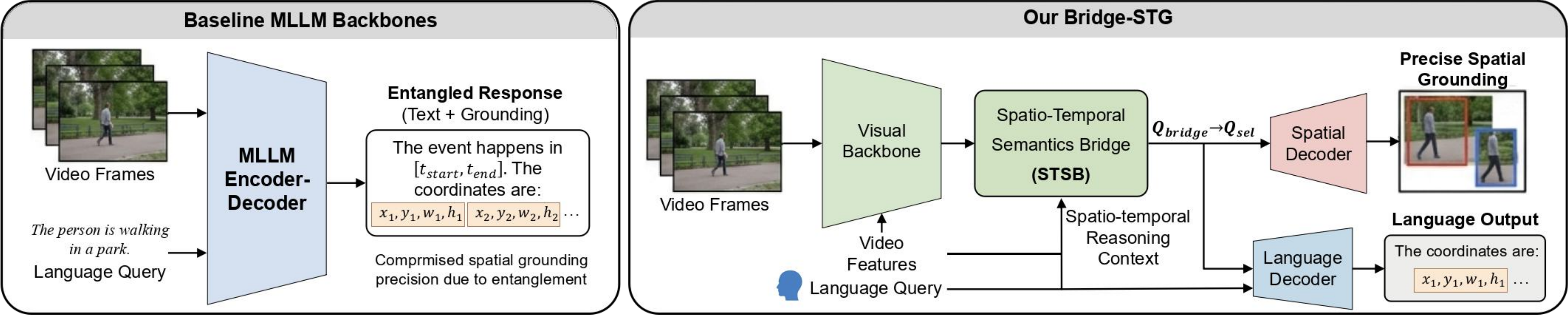}
  \caption{System-level comparison of MLLM-based grounding architectures. Left: Baseline MLLMs generate an entangled response of text and spatial coordinates, which decreases grounding precision. Right: Our Bridge-STG explicitly decouples this process. The STSB distills temporal reasoning context to drive a customized spatial decoder, achieving precise localization.}
  \label{fig:qgsl}
\end{figure*}

\subsection{Decoupled Temporal-Spatial Architecture}
\label{subsec:dtsa}

\textbf{Explicit Temporal Alignment (ETA) Strategy.}
To strengthen the MLLM's event-boundary perception, ETA injects text-formatted timestamps directly into the MLLM's embedding space by appending them after each frame pair's visual tokens. Specifically, for the $i$-th frame pair representing the time interval $[t_i, t_{i+1}]$, we format its corresponding timestamp as a text string $T_i$, which is then tokenized and projected through the MLLM's embedding layer to obtain its token embeddings $\mathbf{e}_{T_i} \in \mathbb{R}^{S \times D}$, where $S$ is the number of tokens produced by the tokenizer and $D$ is the embedding dimension.

The key challenge is that naively appending these token embeddings would disrupt the MLLM's continuous positional embedding space. We address this by assigning each timestamp token embedding a virtual spatial coordinate outside the visual token grid, placing it at position $(W+s, H+s)$ within the temporal slice $i$:
\begin{equation}
    \mathbf{t}'_i = \mathbf{e}_{T_i} + \mathbf{P}(i, W+s, H+s)
\end{equation}
where $s \in \{1,\ldots,S\}$ and $\mathbf{P}$ denotes the positional embedding function. The full input sequence to the MLLM is constructed as:
\begin{equation}
    \mathbf{C}_{full} = \left[\mathbf{v}'_1, \mathbf{t}'_1, \ldots, \mathbf{v}'_{N/2}, \mathbf{t}'_{N/2}\right]
\end{equation}
This preserves the coherence of the spatio-temporal positional embedding space while explicitly anchoring each timestamp to its corresponding visual content.

\textbf{Spatio-Temporal Semantic Bridging Mechanism.}
A key challenge in decoupled STVG is ensuring the spatial grounding module can access the spatio-temporal reasoning context built up during temporal localization. 
To bridge this challenge, STSB introduces a phase transition mechanism that marks the boundary between temporal and spatial reasoning, and extracts the MLLM's accumulated context as a compact semantic representation for downstream use.

Specifically, we extend the MLLM's vocabulary with a transition token \texttt{[DET]}, which the model learns to use once completing its temporal localization response (i.e., after predicting $[t_{start}, t_{end}]$). 
This token serves as an explicit phase boundary, prompting the MLLM to transition from sequence-level temporal reasoning to instance-level spatial grounding. 
Following the \texttt{[DET]} token embedding, a set of $M$ learnable context queries $\mathbf{Q}_{init} \in \mathbb{R}^{M \times D}$ is appended to the input sequence. 
As these queries are processed by the MLLM's layers, they progressively absorb the accumulated spatio-temporal reasoning context encoded in the previous hidden states. 
The final hidden states of these queries are extracted and projected through an MLP to produce the bridging queries $\mathbf{Q}_{bridge} \in \mathbb{R}^{M \times D}$:
\begin{equation}
    \mathbf{Q}_{bridge} = \text{MLP}\left(f_{MLLM}\left( \mathbf{Q}_{init} \mid \mathbf{C}_{full}, \mathbf{T}_{feat} \right) \right)
\end{equation}
where $f_{MLLM}(\cdot)$ denotes the hidden state outputs extracted from the MLLM's last layer. 

Therefore, $\mathbf{Q}_{bridge}$ serves as a semantically condensed representation of both the visual-temporal feature and the linguistic query intent.
This representation provides QGSL with a structured conditional prior for spatial grounding and enables end-to-end gradient flow between the MLLM and the spatial decoder.

\subsection{Query-Guided Spatial Localization Module}
\label{subsec:qgsl}
While $\mathbf{Q}_{bridge}$ provides rich semantic context from MLLM, directly mapping these representations to precise spatial coordinates remains a challenge due to dual-domain visual token redundancy. 
To address this, we propose a Query-Guided Spatial Localization (QGSL) module, a query-conditioned spatial decoder inspired by open-vocabulary detection frameworks~\cite{liu2024grounding,wang2024ov}.

\textbf{Architecture.}
QGSL adopts an encoder-decoder architecture (Fig.~\ref{fig:qgsl}). 
Given an input frame, the image backbone extracts multi-scale visual features, which are processed by an $n$-layer image encoder to produce hierarchical feature representations $\{\mathbf{E}^{l}_{img}\}_{l=1}^{n}$ ($n=6$). 
Additionally, we remove the text backbone and instead condition the entire text decoding process on $\mathbf{Q}_{bridge}$ from STSB.
This enables the decoder to directly use the MLLM's accumulated spatio-temporal reasoning context rather than text embeddings.

\begin{figure}
  \centering
  \includegraphics[width=\linewidth]{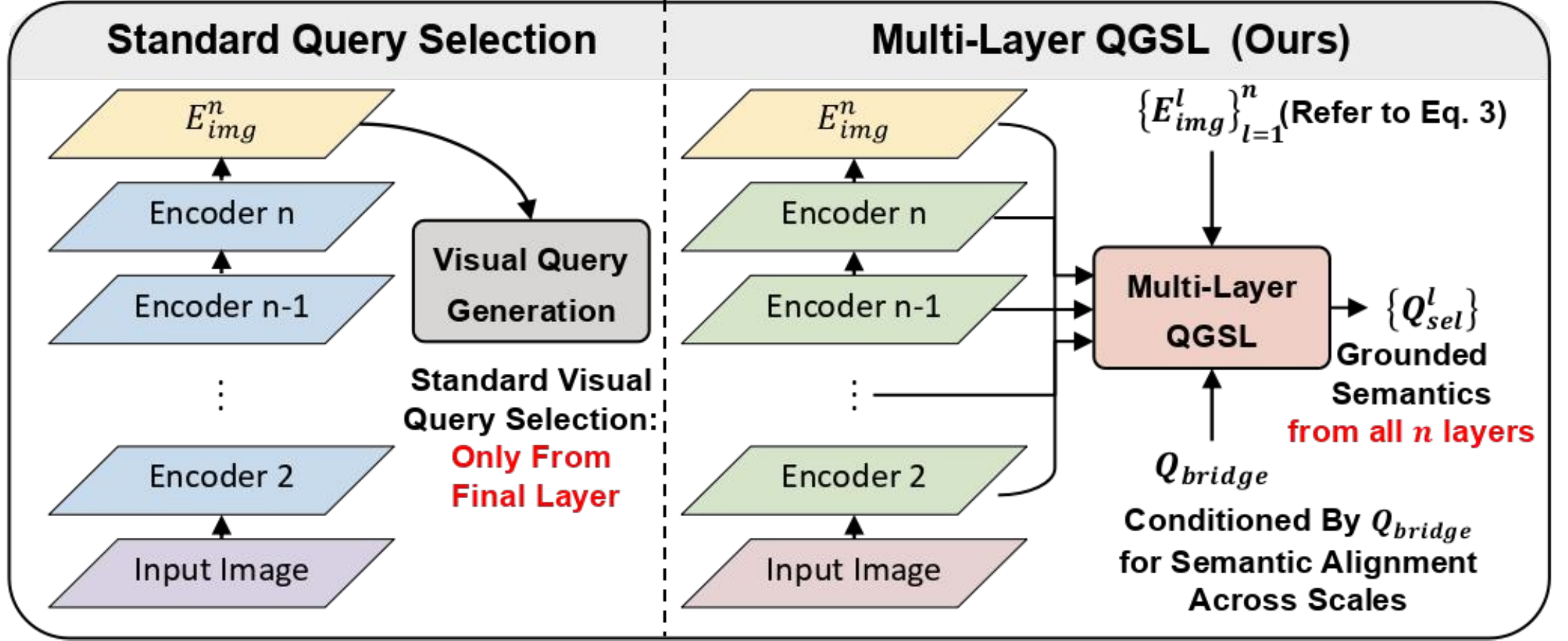} 
  \caption{The comparison of visual query selection. Left: Standard methods extract visual features ($E_{img}^n$) only from the final encoder layer. Right: Our Multi-Layer QGSL aggregates features ($\{E_{img}^l\}_{l=1}^n$) from all $n$ layers. Conditioned directly by $Q_{bridge}$, it produces semantically aligned spatial queries ($\{Q_{sel}^l\}$) to robustly capture grounded semantics.}
  \label{fig:multi-layer}
\end{figure}

\textbf{Multi-Layer Interactive Queries.}
As shown in Fig.~\ref{fig:multi-layer}, standard query selection selects candidate queries solely from the last encoder layer, which may miss fine-grained spatial features encoded in intermediate layers.
To overcome this limitation, we introduce multi-layer interactive queries that aggregate candidate features across all $n$ encoder layers.
Specifically, from each encoder layer $l$, we select the top-$K$ image features most relevant to $\mathbf{Q}_{bridge}$ based on cosine similarity (top-$K = 900$ in our experiments):
\begin{equation}
    \mathbf{Q}^{l}_{sel} = \mathrm{TopK}\left(\mathbf{E}^{l}_{img}, \mathbf{Q}_{bridge}\right), \quad l = 1, \ldots, n
\end{equation}
In total, we get $K \times n$ candidate queries: $\mathbf{Q}_{sel} = \{\mathbf{Q}^{l}_{sel}\}_{l=1}^{n}$. 
The selected multi-layer queries $\mathbf{Q}_{sel}$ serve as the initialized object queries to the spatial decoder, which performs cross-attention with the image encoder features to produce the final bounding box predictions:
\begin{equation}
    \hat{\mathbf{B}} = \mathcal{D}\left(\mathbf{E}_{img}, 
    \mathbf{Q}_{sel}\right)
\end{equation}
where $\mathcal{D}$ and $\mathbf{E}_{img}$ denote spatial decoder and encoded image features, respectively. 
By expanding the candidate pool, multi-layer interactive queries enrich spatial feature diversity, particularly benefiting localization of small or occluded objects in video data.

\textbf{Image-Query Alignment.}
Since query selection relies on image-query similarity, it is essential that $\mathbf{Q}_{bridge}$ and the selected image features are semantically aligned in a shared embedding space. 
A simple approach is to optimize the positive cosine similarity; however, this lacks the discriminative penalty needed to separate the target from visually similar background distractors, which often leads to feature collapse.

To ensure the selected image features are target-aware and robust against redundant visual tokens, we formulate the image-query alignment as a contrastive learning objective.
Specifically, we treat the Top-$K$ selected multi-layer queries as positive samples while uniformly sampling unselected visual tokens as negative samples. The InfoNCE-based alignment loss is defined as:
\begin{equation}
  \mathcal{L}_{align} = - \frac{1}{Kn} \sum_{l=1}^{n} \sum_{j=1}^{K} \log
  \frac{e^{\cos(Q_{sel}^{l,j},\, \tilde{Q}_{bridge}) / \tau}}
       {e^{\cos(Q_{sel}^{l,j},\, \tilde{Q}_{bridge}) / \tau} + \displaystyle\sum_{v \in \mathcal{N}^l} e^{\cos(v,\, \tilde{Q}_{bridge}) / \tau}}
  \end{equation}      
where $Q_{sel}^{l,j}$ denotes the $j$-th selected query from layer $l$, $\tilde{Q}_{bridge} = \frac{1}{M}\sum_{m=1}^{M}Q_{bridge}^{m}$ is the mean vector of the $M$ bridging queries, and $cos(\cdot,\cdot)$ represents the cosine similarity function. $\mathcal{N}^l$ is the set of negative visual features sampled from the remaining unselected tokens in layer $l$, and $\tau$ is a learnable temperature hyperparameter.
$|\mathcal{N}^l|$ negative features are uniformly sampled from the remaining unselected tokens in layer $l$ for each positive pair, and the cosine similarity serves as the relevance metric for both positive and negative comparisons.
The mean pooling over $M$ bridging queries provides a compact and stable semantic anchor for contrastive alignment,  where the grounding target is a single referred object.
This loss supervises the query selection process, ensuring that the selected image features are visually discriminative and semantically coherent with the grounding query.

\begin{figure}
  \centering
  \includegraphics[width=1\linewidth]{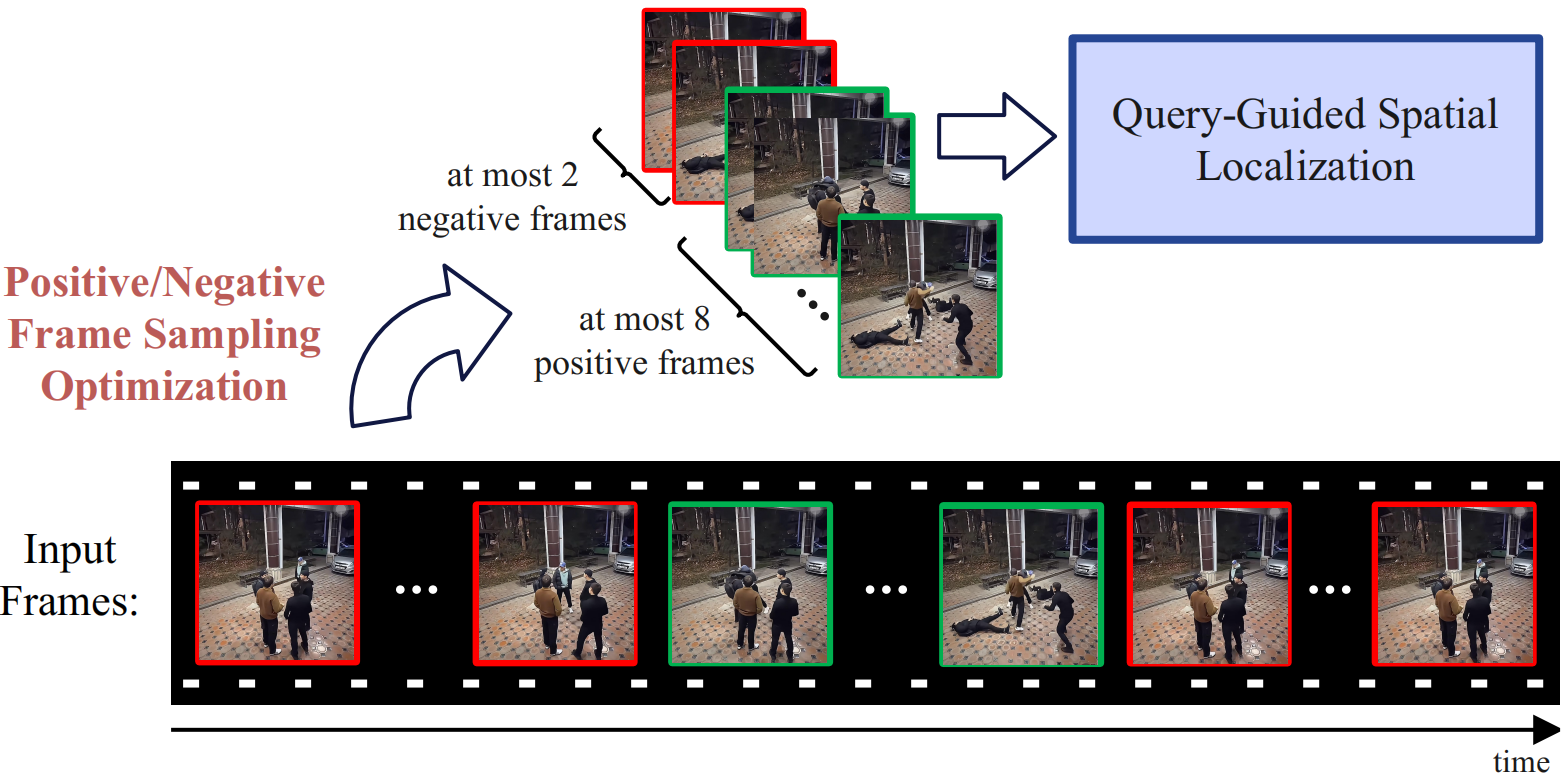} 
  \caption{The positive/negative frame sampling strategy.}
  \label{fig:pn_frame_sampling}
\end{figure}

\textbf{Positive/Negative Frame Sampling.}
During training, feeding all video frames into QGSL would face training-time frame redundancy and risk out-of-memory issues for long videos.
We address this with a positive/negative (P/N) frame sampling strategy (Fig.~\ref{fig:pn_frame_sampling}). 

Given a video with $N$ total frames, $K$ of which are in the ground-truth temporal window $[t_{start}^{gt}, t_{end}^{gt}]$ (positive frames), we randomly sample up to $N_p$ frames from the positives and up to $N_n$ frames from the remaining $N-K$ negative frames. 
In practice, we set $N_p = 8$ and $N_n = 2$, yielding at most 10 frames per training iteration.
The inclusion of negative frames enables the QGSL to learn to distinguish the target object from visually similar background regions. 
During inference, all frames within $[t_{start}^{predict}, t_{end}^{predict}]$ are passed to QGSL for spatial localization without sampling constraints.

\subsection{Overall Training Objectives}
\label{subsec:overall_loss}
Bridge-STG is trained end-to-end via supervised fine-tuning with a joint objective that simultaneously supervises temporal localization and spatial grounding:
\begin{equation}
    \mathcal{L} = \lambda_1 \mathcal{L}_{token} + \lambda_2 \mathcal{L}_{spatial}
\end{equation}

$\mathcal{L}_{token}$ is the standard autoregressive cross-entropy loss over the MLLM's output token sequence, supervising both the predicted temporal window $[t_{start}, t_{end}]$ and the \texttt{[DET]} transition token. $\mathcal{L}_{spatial}$ is the spatial grounding loss of QGSL, augmented with our image-query alignment term:
\begin{equation}
    \mathcal{L}_{spatial} = \alpha \mathcal{L}_{obj} + \beta\mathcal{L}_{box} + \gamma\mathcal{L}_{giou} + \delta\mathcal{L}_{dn} + \eta\mathcal{L}_{align}
\end{equation}

where $\mathcal{L}_{obj}$ is a binary objectness loss that supervises whether each decoded query corresponds to the target object (foreground) or not (background).
$\mathcal{L}_{box}$ and $\mathcal{L}_{giou}$ are the L1 bounding box regression loss and Generalized IoU loss, respectively, which are complementary in supervising spatial precision—$\mathcal{L}_{box}$ provides direct coordinate supervision, while $\mathcal{L}_{giou}$ optimizes overlap quality in a scale-invariant manner.
$\mathcal{L}_{align}$ is the image-query alignment loss introduced in Sec.~\ref{subsec:qgsl}.

Furthermore, to accelerate bipartite matching convergence and stabilize spatial decoder training, we incorporate a contrastive denoising loss $\mathcal{L}_{dn}$:
\begin{equation}
    \mathcal{L}_{dn} = \lambda_{cls\_dn}\mathcal{L}_{cls\_dn} + \lambda_{box\_dn}\mathcal{L}_{box\_dn} + \lambda_{giou\_dn}\mathcal{L}_{giou\_dn}
\end{equation} 
$\mathcal{L}_{dn}$ is the denoising loss that feeds perturbed ground-truth boxes as auxiliary queries during training to stabilize decoder convergence and is removed at inference time.
Specifically, it consists of a classification reconstruction loss ($\mathcal{L}_{cls\_dn}$), an L1 box regression loss ($\mathcal{L}_{box\_dn}$), and a GIoU loss ($\mathcal{L}_{giou\_dn}$).

\begin{table*}[t]
\scriptsize
\centering
\caption{The overview of data source.}
\label{tab:dataset}
    \begin{tabular}{p{3cm} c c c}
    \hline
    \textbf{Training Stage} &  \textbf{Task} & \textbf{Data Source} & \textbf{\# of Samples} \\
    \hline
    \multirow{5}{*}{\makecell[c]{Multi-Task\\Instruction Tuning\\}} 
      & Spatial-Temporal Video Grounding 
      & HCSTVG-v1\&v2~\cite{tang2021human}, VidSTG~\cite{zhang2020does}, Self-collected 
      & 127K \\
      & Video Temporal Grounding 
      & Charades-STA~\cite{gao2017tall} 
      & 12K \\
      & Video Object Tracking 
      & GOT-10k~\cite{huang2019got} 
      & 10K \\
      &{Video Question Answering}
      &{NextQA~\cite{xiao2021next}, Clevrer~\cite{yi2019clevrer}}
      &{90K}\\
      & Referring Expression Comprehension 
      & RefCOCO~\cite{kazemzadeh2014referitgame}, RefCOCO+~\cite{kazemzadeh2014referitgame}, RefCOCOg~\cite{mao2016generation} 
      & 120K \\
    \hline
    \end{tabular}
\end{table*}

\begin{table*}[t]
\scriptsize
\centering
\setlength{\tabcolsep}{12pt}
\caption{Comparison with existing state-of-the-art models on VidSTG~\cite{zhang2020does} test set (\%).}
\label{tab:vidstg_res}
\begin{tabularx}{0.9\textwidth}{>{\raggedright\arraybackslash}Xcccccccc}
\hline 
\multirow{2}{*}{\textbf{Model}}&\multicolumn{4}{c}{\textbf{{Declarative Sentences}}} & 
\multicolumn{4}{c}{\textbf{{Interrogative Sentences}}}\\
\cmidrule(lr){2-5} \cmidrule(lr){6-9} 
& \text{m\_tIoU} & \text{m\_vIoU} & \text{vIoU@0.3} & \text{vIoU@0.5} & \text{m\_tIoU} & \text{m\_vIoU} & \text{vIoU@0.3} & \text{vIoU@0.5} \\
\hline
\multicolumn{9}{c}{\textit{\textcolor[gray]{0.5}{Non-generative and task-specific models}}} \\
TubeDETR~\cite{yang2022tubedetr} & 48.1 & 30.4 & 42.5 & 28.2 & 46.9 & 25.7 & 35.7 & 23.2 \\
CG-STVG~\cite{gu2024context}  & 51.4 & 34.0 & 47.7 & 33.1 & 49.9 & 29.0 & 40.5 & 27.5 \\
TA-STVG~\cite{gu2025knowing} & {51.7} & 34.4 & 48.2 & 33.5 & \textbf{50.2} & 29.5 & 41.5 & 28.0 \\
\hline
\multicolumn{9}{c}{\textit{\textcolor[gray]{0.5}{7B-based MLLMs}}} \\
LLaVA-ST~\cite{li2025llava} & 44.1 & 14.2 & 18.5 & 7.5 & 42.9 & 11.5 & 14.3 & 5.9 \\
VideoMolmo~\cite{ahmad2025videomolmo} & 41.7 & 15.6 & - & - & 30.2 & 11.7 & 15.2 & 7.3 \\
SpaceVLLM~\cite{wang2025spacevllm} & 47.7 & 27.4 & 39.1 & 26.2 & 48.5 & 25.4 & 35.9 & 22.2 \\
\hline
\rowcolor{gray!20}
\textbf{Bridge-STG} & \textbf{52.6} & \textbf{37.2} & \textbf{52.4} & \textbf{37.4} & {50.1} & \textbf{31.3} & \textbf{43.8} & \textbf{31.2} \\
\hline
\end{tabularx}
\end{table*}

\section{Experiments}
\subsection{Experimental Settings}
\textbf{Implementation Details.} We employ Qwen3-VL-8B~\cite{bai2025qwen3} as the pre-trained MLLM, optimized via AdamW~\cite{loshchilov2017decoupled} ($lr=1e-4$, weight decay $=0$, cosine scheduler with 0.1 warmup ratio).
We apply LoRA~\cite{hu2022lora} ($r=8, \alpha=32$) with a batch size of 32.
The number of bridging queries is 8. 
Loss weights are $\lambda_1=1.0, \lambda_2=0.02$, with $\alpha=1.0, \beta=0.5, \gamma=2.0, \delta=1.0, \eta=1.0$. 
Within the denoising branch, the internal weights for $\lambda_{cls\_dn}, \lambda_{box\_dn}$, and $\lambda_{giou\_dn}$ are set to 1.0, 5.0, and 2.0, respectively. 
Videos are uniformly sampled at 2 FPS. 
QGSL processes 10 frames (8 positive, 2 negative) per iteration.
Bridge-STG is trained on 8 NVIDIA H100 GPUs for 16.4 hours. 

\begin{table} 
\centering
\scriptsize 
\setlength{\tabcolsep}{8pt}
\caption{Comparison with STVG methods on HCSTVG~\cite{tang2021human}. }
\label{tab:hcstvgv2_res}
\begin{tabularx}{\linewidth}{>{\raggedright\arraybackslash}Xcccccccc}
\toprule
\textbf{Model} & \textbf{m\_tIoU} & \textbf{m\_vIoU} & \textbf{vIoU@0.3} & \textbf{vIoU@0.5} \\
\hline
\multicolumn{5}{c}{\textit{\textcolor[gray]{0.5}{Non-generative and task-specific models}}} \\
TubeDETR~\cite{yang2022tubedetr} & 53.96 & 36.4 & 58.8 & 30.6 \\
CG-STVG~\cite{gu2024context}  & 60.0 & 39.5 & 64.5 & 36.3 \\
TA-STVG~\cite{gu2025knowing} & {60.4} & {40.2} & {65.8} & {36.7} \\
\hline
\multicolumn{5}{c}{\textit{\textcolor[gray]{0.5}{7B-based MLLMs}}} \\
LLaVA-ST~\cite{li2025llava}  & 21.2 & 7.6 & 4.3 & 0.6 \\
VideoMolmo~\cite{ahmad2025videomolmo} & 44.6 & 26.8 & 37.7 & 12.4 \\
SpaceVLLM~\cite{wang2025spacevllm} & 58.0 & 34.0 & 56.9 & 24.7 \\
\hline
\rowcolor{gray!20}
\textbf{Bridge-STG} & \textbf{64.1} & \textbf{41.5} & \textbf{67.5} & \textbf{38.6} \\
\bottomrule
\end{tabularx}
\end{table}

 \noindent\textbf{Training Datasets.}
Following~\cite{li2025llava, wang2025spacevllm}, three existing STVG benchmarks (HCSTVG-v1\&v2 ($\sim$107K)~\cite{tang2021human} and VidSTG ($\sim$10K)~\cite{zhang2020does}) are used for training Bridge-STG to enhance spatio-temporal understanding capacity. Additionally, a synthetic dataset generated from the ReVOS dataset~\cite{yan2024visa} acts as the augmented STVG training data ($\sim$10K). 
To prevent overfitting to limited spatial-temporal patterns, we introduce multi-task instruction tuning using VTG, VOT, REC, and VQA datasets.
The details are shown in Tab.~\ref{tab:dataset}.

\noindent\textbf{Evaluation Datasets.}
We use 10 benchmarks to perform a comprehensive evaluation for Bridge-STG, covering STVG, VTG, VOT, REC and VQA.
Following~\cite{
gu2024context, gu2025knowing, su2021stvgbert, yang2022tubedetr}, the evaluation is first given on two standard STVG benchmarks, HC-STVG~\cite{tang2021human} and VidSTG~\cite{zhang2020does}. To evaluate cross-task transfer, we adopt Charades-STA~\cite{gao2017tall} for VTG and 
GOT-10K~\cite{huang2019got} for VOT. For REC, the RefCOCO \cite{kazemzadeh2014referitgame, mao2016generation} is used. 
We additionally report results on VideoMME~\cite{fu2025video} for VQA.

\noindent \textbf{Baselines.}
For STVG, we compare two types of models: 1) MLLMs (7B) including Qwen2.5-VL \cite{bai2025qwen2}, SpaceVLLM \cite{wang2025spacevllm}, LLaVA-ST \cite{li2025llava}, and VideoMolmo \cite{ahmad2025videomolmo}.
All models are evaluated using their officially released checkpoints fine-tuned on STVG instruction data.
2) Non-generative task-specific models, including TubeDETR \cite{yang2022tubedetr}, CG-STVG \cite{gu2024context}, and TA-STVG \cite{gu2025knowing}.
These models serve as SOTA baselines representing traditional methods. To validate generalization performance, we compare Bridge-STG with currently high-performing models including: TimeSuit \cite{zeng2024timesuite}, VLG-LLM \cite{guo2025vtg}, HawkEye \cite{wang2024hawkeye}, EaTR \cite{jang2023knowing}, and QD-DETR \cite{moon2023query} for VTG; AQATrack \cite{xie2024autoregressive}, DuTrack \cite{li2025dynamic}, R1-Track \cite{wang2025r1}, and ReasoningTrack \cite{wang2025reasoningtrack} for VOT; G-DINO \cite{liu2024grounding}, GLEE \cite{wu2024general}, Elysium \cite{wang2024elysium}, Qwen2.5-VL \cite{bai2025qwen2} for REC;  
VideoLLaVA \cite{lin2024video}, Videollama2.1 \cite{cheng2024videollama} and LLaVA-OV \cite{li2024llava} for VQA.

\noindent\textbf{Evaluation Metrics.}
Following~\cite{gu2024context,gu2025knowing, jin2022embracing,su2021stvgbert, yang2022tubedetr}, m\_tIoU, m\_vIoU, vIoU@R are adopted as evaluation metrics for STVG. m\_tIoU reports the mean temporal Intersection-over-Union (tIoU) of predicted versus ground-truth intervals, evaluating temporal grounding. m\_vIoU computes the average 3D IoU of spatio-temporal tubes to assess spatial grounding. vIoU@R measures the proportion of samples with vIoU exceeding a threshold R (e.g., 0.3, 0.5), indicating performance under precise localization demands. Additionally, we report recall at varying IoU thresholds for VTG. AO (Average Overlap) and SR (Success Rate) described in~\cite{huang2019got} are used for VOT.
For REC and VQA, we use IoU@0.5 and standard accuracy, respectively.

\subsection{Performance on STVG}
\label{subsec:exp_stvg}

\begin{table}
\centering
\scriptsize 
\setlength{\tabcolsep}{20pt}
\caption{Results on Charades-STA~\cite{gao2017tall} for VTG task.}
\label{tab:vtg_res}
\begin{tabularx}{\linewidth}{>{\raggedright\arraybackslash}Xcc}
\toprule
\textbf{Model} & \textbf{R@1\textsubscript{IoU=0.5}} & \textbf{R@1\textsubscript{IoU=0.7}} \\
\hline
\multicolumn{3}{c}{\textit{\textcolor[gray]{0.5}{Non-generative and task-specific models (Spatio-Temporal)}}}\\
CG-STVG~\cite{gu2024context} & 20.0 & 7.1 \\
TA-STVG~\cite{gu2025knowing}  & 16.3 & 5.2\\
\multicolumn{3}{c}{\textit{\textcolor[gray]{0.5}{Non-generative and task-specific models (only-Temporal)}}} \\
QD-DETR~\cite{moon2023query}& 57.3 & 32.6 \\
EaTR~\cite{jang2023knowing} & 68.4 & 44.9 \\
\hline
\multicolumn{3}{c}{\textit{\textcolor[gray]{0.5}{7B-based MLLMs (only-Temporal)}}} \\
VTG-LLM~\cite{guo2025vtg}& 57.2 & 33.4 \\
HawkEye~\cite{wang2024hawkeye}& 58.3 & 28.8 \\
TimeSuite~\cite{zeng2024timesuite} & 67.1 & 43.0 \\
\multicolumn{3}{c}{\textit{\textcolor[gray]{0.5}{7B-based MLLMs (Spatio-Temporal)}}}\\
LLaVA-ST~\cite{li2025llava} & 44.8 & 23.4 \\
SpaceVLLM~\cite{wang2025spacevllm} & 63.6 & 38.5 \\
\hline
\rowcolor{gray!20}
\textbf{Bridge-STG} & \textbf{70.3} & \textbf{49.3}\\
\bottomrule
\end{tabularx}
\end{table}

\begin{table}
\centering
\scriptsize 
\setlength{\tabcolsep}{16pt}
\caption{Results on GOT-10K~\cite{huang2019got} for VOT task.}
\label{tab:vot_res}
\begin{tabularx}{\linewidth}{>{\raggedright\arraybackslash}Xccc}
\toprule
\textbf{Model} & \textbf{AO} & \textbf{SR@0.5} & \textbf{SR@0.7} \\
\hline
\multicolumn{4}{c}{\textit{\textcolor[gray]{0.5}{Non-generative and task-specific models}}} \\
AQATrack~\cite{xie2024autoregressive} & 76.0 & 85.2 & 74.9 \\
DuTrack~\cite{li2025dynamic} & {77.8} & 88.2 & 76.0 \\
\hline
\multicolumn{4}{c}{\textit{\textcolor[gray]{0.5}{7B-based MLLMs (only-Spatial)}}} \\
R1-Track~\cite{wang2025r1}& 68.0 & 76.6 & 63.7 \\
ReasoningTrack~\cite{wang2025reasoningtrack} & {77.8} & {88.5} & {77.0}\\
\hline
\rowcolor{gray!20}
\textbf{Bridge-STG} & \textbf{79.3} & \textbf{88.8} & \textbf{78.1} \\
\bottomrule
\end{tabularx}
\end{table}

\noindent\textbf{VidSTG.}  
We first evaluate the proposed method on the challenging VidSTG, which contains both declarative and interrogative sentences. 
As shown in Tab.~\ref{tab:vidstg_res}, MLLM-based methods like LLaVA-ST and VideoMolmo struggle with complex scenarios, achieving low m\_vIoU scores of only 14.2 and 15.6 on declarative sentences, respectively. 
Their performance further degrades on interrogative sentences (dropping to 11.5 and 11.7 m\_vIoU), which require implicit target reasoning based on video content.

In contrast, our Bridge-STG achieves state-of-the-art performance among generative models, significantly outperforming the latest MLLM baseline SpaceVLLM, by $+9.8$ in m\_vIoU for declarative sentences (37.2 vs. 27.4) and $+5.9$ for interrogative sentences (31.3 vs. 25.4). 
Furthermore, Bridge-STG bridges the performance gap with non-generative task-specific models.
Our method achieves the best performance on all metrics for declarative sentences (e.g., beating TA-STVG by 2.8 in m\_vIoU).
For interrogative sentences, Bridge-STG achieves better spatial grounding (31.3 vs. 29.5 m\_vIoU) while maintaining competitive temporal localization (50.1 vs. 50.2 m\_tIoU) compared to TA-STVG.
Overall, Bridge-STG achieves an average m\_vIoU of 34.3 across both declarative and interrogative subsets, compared to 26.4 for SpaceVLLM.
This highlights the robustness of our decoupled architecture in accurately extracting spatio-temporal features under complex linguistic queries.

\noindent\textbf{HCSTVG-v2.} 
The results in Tab.~\ref{tab:hcstvgv2_res} demonstrate the superiority of Bridge-STG on the declarative-only HCSTVG benchmark.
Compared to other MLLM-based methods, Bridge-STG achieves 64.1 m\_tIoU and 41.5 m\_vIoU, outperforming SpaceVLLM by 6.1 and 7.5, respectively. 
The improvement is particularly substantial under strict spatial evaluation, yielding a $+13.9$ on vIoU@0.5 (38.6 vs. 24.7). 
Furthermore, compared to LLaVA-ST, our method yields improvements of 42.9 and 33.9 on m\_tIoU and m\_vIoU, respectively.

Remarkably, Bridge-STG comprehensively surpasses the current best task-specific model, TA-STVG, in all four evaluation metrics (e.g., $+3.7$ in m\_tIoU and $+1.3$ in m\_vIoU). 
Considering that task-specific methods rely on customized region-proposal designs and proprietary dataset optimizations, our method maintains the MLLMs' generalization capabilities on video understanding.
This validates the effectiveness of STSB mechanism and QGSL module.

\begin{table}[t]
\scriptsize
\setlength{\tabcolsep}{4pt}
\caption{The IoU@0.5 performance on RefCOCO~\cite{kazemzadeh2014referitgame}, RefCOCO+~\cite{kazemzadeh2014referitgame}
and RefCOCOg~\cite{mao2016generation} for REC task.}
\label{tab:rec_res}
\begin{tabularx}{\linewidth}{>{\raggedright\arraybackslash}X|ccc|ccc|cc}
\toprule
\multirow{2}{*}{\textbf{Model}}&\multicolumn{3}{c}{\textbf{RefCOCO}} & 
\multicolumn{3}{c}{\textbf{RefCOCO+}} &
\multicolumn{2}{c}{\textbf{RefCOCOg}} \\
\cmidrule(lr){2-4} \cmidrule(lr){5-7} \cmidrule(lr){8-9} 
\addlinespace[-2pt]
&val & test-A & test-B & val & test-A & test-B & val & test \\
\hline
\multicolumn{9}{c}{\textit{\textcolor[gray]{0.5}{Non-generative and task-specific models}}} \\
G-DINO~\cite{liu2024grounding} & 90.6 & 93.2 & 88.2 & 88.2 & 89.0 & 75.9 & 86.1 & 87 \\
GLEE~\cite{wu2024general} & 91.0 & - & - & 86.4 & - & - & 82.6 & - \\
\hline
\multicolumn{9}{c}{\textit{\textcolor[gray]{0.5}{7B-based MLLMs (only-Spatial)}}} \\
Elysium~\cite{wang2024elysium} & 89.1 & 92.1 & 85.0 & 82.9 & 88.9 & 75.6 & 82.9 & 83.6 \\
Qwen2.5VL~\cite{bai2025qwen2}& 90.0 & 92.5 & 85.4 & 84.2 & 89.1 & 76.9 & 87.2 & 87.2 \\
\hline
\multicolumn{9}{c}{\textit{\textcolor[gray]{0.5}{7B-based MLLMs (Spatial-Temporal)}}} \\
LLaVA-ST~\cite{li2025llava} & 90.1 & 93.2 & 85.0 & 86.0 & 91.3 & 78.8 & 86.7 & 87.4 \\
SpaceVLLM~\cite{wang2025spacevllm} & 90.8 & 93.4 & 87.0 & 86.3 & 90.9 & 79.8 & 86.8 & 88.0 \\
\hline
\rowcolor{gray!20}
\textbf{Bridge-STG} &\textbf{91.9} & \textbf{94.6} & \textbf{89.0} & \textbf{87.8} & \textbf{91.9} & \textbf{82.3} & \textbf{89.3} & \textbf{89.5} \\
\bottomrule
\end{tabularx}
\end{table}

\subsection{Performance on Cross-Task Transfer}
\label{subsec:exp_gen}
We further evaluate the cross-task transfer capability of Bridge-STG on four video understanding tasks (VTG, VOT, REC, and VQA). 

\noindent\textbf{Video Temporal Grounding (VTG).} Tab.~\ref{tab:vtg_res} presents the performance of our model on Charades-STA \cite{gao2017tall} for VTG. 
A critical observation is that traditional task-specific STVG models (e.g., CG-STVG, TA-STVG) exhibit severe performance degradation when generalized to pure temporal grounding, scoring only 20.0 and 16.3 on R@1 (IoU=0.5).
This indicates that their architectures overfit to specific dataset patterns.
While existing spatio-temporal MLLMs like SpaceVLLM demonstrate better generalization (63.6 at IoU=0.5), they still underperform compared to dedicated temporal models.

In contrast, Bridge-STG outperforms the latest spatio-temporal baseline, SpaceVLLM, by significant improvements of $+6.7$ and $+10.8$ on R@1\textsubscript{IoU=0.5} and R@1\textsubscript{IoU=0.7}, respectively.
Our method also surpasses models specifically optimized for the VTG task, including the leading temporal-only MLLM (TimeSuite, +3.2 on R@1\textsubscript{IoU=0.5}) and the task-specific VTG model (EaTR, +1.9 on R@1\textsubscript{IoU=0.5}). 
This generalization capacity stems from our temporal anchoring and decoupled design, which prevents task-overfitting and ensures robust temporal perception even in out-of-domain scenarios.

\noindent\textbf{Video Object Tracking (VOT).} 
To evaluate the fine-grained tracking consistency of Bridge-STG, we extend our evaluation to the single-object tracking task. 
As shown in Tab.~\ref{tab:vot_res}, our method surpasses the tracking-specific MLLM baseline ReasoningTrack, achieving an AO of 79.3.
Furthermore, Bridge-STG demonstrates robustness in precise bounding box estimation, achieving 88.8 and 78.1 on SR@0.5 and SR@0.7, respectively.
Unlike R1-Track or ReasoningTrack, which use tracking-specific heads or RL tracking pipelines for VOT, our STSB and multi-layer QGSL modules effectively preserve instance-level temporal consistency and fine-grained spatial representations, translating generalized video grounding capabilities into highly accurate object tracking.
We follow the GOT-10K one-shot protocol~\cite{huang2019got} with strict train-test class separation.

\noindent\textbf{Referring Expression Comprehension (REC).}
To assess fine-grained spatial understanding capability, we evaluate Bridge-STG on REC benchmarks: RefCOCO, RefCOCO+, and RefCOCOg. 
As shown in Tab.~\ref{tab:rec_res}, our model achieves state-of-the-art performance on 8 evaluation metrics across the three datasets.
Specifically, Bridge-STG surpasses spatial-only MLLMs (e.g., Elysium) and specialized non-generative grounding models (e.g., G-DINO). 
The experiment results show that our Query-Guided Spatial Localization (QGSL) module, combined with the contrastive image-query alignment, effectively preserves and enhances fine-grained spatial semantics.

\begin{table}[t]
\centering
\scriptsize 
\setlength{\tabcolsep}{20pt}
\caption{Results on VideoMME~\cite{fu2025video} for VQA task.}
\label{tab:vqa_res}
\begin{tabularx}{0.9\linewidth}{>{\raggedright\arraybackslash}Xcc}
\toprule
\textbf{Model} & \textbf{w/o subs} & \textbf{w/ subs} \\
\hline
\multicolumn{3}{c}{\textit{\textcolor[gray]{0.5}{7B-based MLLMs}}} \\
Video-LLaVA~\cite{lin2024video} & 39.9 & 41.6 \\
Videollama2.1~\cite{cheng2024videollama} & 54.9 & 56.4 \\
LLaVA-OV~\cite{li2024llava} & 58.2 & 61.5 \\
SpaceVLLM~\cite{wang2025spacevllm} & 60.0 & 65.6 \\
\hline
\rowcolor{gray!20}
\textbf{Bridge-STG} & \textbf{67.9} & \textbf{74.8} \\
\bottomrule
\end{tabularx}
\end{table}

\noindent\textbf{Video Question Answering (VQA).}
To assess general video comprehension, we evaluate Bridge-STG on the Video-MME. 
As shown in Tab.~\ref{tab:vqa_res}, our model shows robust reasoning capabilities, achieving 67.9\% without subtitles and 74.8\% with subtitles.
Notably, Bridge-STG outperforms the recent spatio-temporal baseline SpaceVLLM.

\begin{table}[!t]
\centering
\scriptsize
\setlength{\tabcolsep}{8pt}
\caption{Results of ablation studies on VidSTG~\cite{zhang2020does}.}
\label{tab:ablation_all}
\begin{tabularx}{\linewidth}{>{\raggedright\arraybackslash}Xcccc}
\toprule
\textbf{Ablation Setting} &\textbf{ m\_tIoU} & \textbf{m\_vIoU} & \textbf{vIoU@0.3} & \textbf{vIoU@0.5}\\
\hline
\multicolumn{5}{c}{\textit{\textcolor[gray]{0.5}{Model Architecture \& Training Strategy}}} \\
w/o ETA & 50.2 & 32.3 & 44.9 & 30.8 \\
w/o STSB & 50.5 & 32.6 & 46.0 & 31.3 \\
w/o P/N-Frame & 51.1 & 34.7 & 48.0 & 33.7 \\
\rowcolor{gray!20}
\textbf{Bridge-STG (Default)} & \textbf{52.6} & \textbf{37.2} & \textbf{52.4} & \textbf{37.4} \\
\hline
\multicolumn{5}{c}{\textit{\textcolor[gray]{0.5}{Number of Bridging Queries}}} \\
0 Queries & 50.1 & 32.6 & 46.0 & 31.3 \\
16 Queries & 51.8 & 36.7 & 50.9 & 36.1 \\
\rowcolor{gray!20}
\textbf{8 Queries (Default)} & \textbf{52.6} & \textbf{37.2} & \textbf{52.4} & \textbf{37.4} \\
\hline
\multicolumn{5}{c}{\textit{\textcolor[gray]{0.5}{Positive-Negative Frame Ratio}}} \\
10 : 0 & 51.1 & 34.7 & 48.0 & 33.7 \\
5 : 5 & 52.0 & 36.9 & 52.1 & 36.8 \\
2 : 8 & 51.4 & 34.9 & 49.6 & 34.3 \\
\rowcolor{gray!20}
\textbf{8 : 2 (Default)} & \textbf{52.6} & \textbf{37.2} & \textbf{52.4} & \textbf{37.4} \\
\hline
\multicolumn{5}{c}{\textit{\textcolor[gray]{0.5}{Loss Weight Coefficients ($\lambda_1 : \lambda_2$)}}} \\
1 : 1 & 50.3 & 37.0 & 51.1 & 37.1 \\
\rowcolor{gray!20}
\textbf{1 : 0.02 (Default)} & \textbf{52.6} & \textbf{37.2} & \textbf{52.4} & \textbf{37.4} \\
\bottomrule
\end{tabularx}
\end{table}

\subsection{Ablation Study}
In this section, we conduct ablation studies to validate the core components of Bridge-STG. 
All ablation experiments are evaluated on the declarative sentences subset of the VidSTG benchmark.

\noindent \textbf{Model Architecture}.
Tab.~\ref{tab:ablation_all} illustrates the importance of our architectural designs.
Removing ETA (w/o ETA) results in performance degradation in m\_tIoU dropping by 2.4 and m\_vIoU dropping from 37.2 to 32.3.
This confirms that without explicit textual timestamp injections, the MLLM loses its structured temporal anchoring, directly decreasing both event boundary perception and subsequent spatial localization. 
Similarly, excluding STSB (w/o STSB) leads to a substantial decrease across all metrics.
Without STSB, the spatial decoder is isolated from the MLLM's sequence-level temporal reasoning context, proving that our learnable bridging queries are essential for maintaining cross-module semantic coherence.

\noindent\textbf{Training Strategy.}
 As shown in Tab.~\ref{tab:ablation_all}, removing P/N-Frame strategy significantly decreases the model's spatial performance, with m\_vIoU dropping by 2.5 (37.2 vs. 34.7).
 This shows that intentionally training on negative frames forces the model to learn discriminative features against visually similar background distractors.

\noindent\textbf{Number of Bridging Queries}.
Tab.~\ref{tab:ablation_all} presents performance under different numbers of bridging queries.
Using 0 queries represents a complete disconnect, yielding the lowest performance. 
Further increasing from 8 to 16 queries yields a slight performance decrease (m\_tIoU drops to 51.8), because excessive queries introduce redundant noise that disrupts the spatial decoder's cross-attention.

\noindent\textbf{Hyperparameter.} 
We analyze the robustness of hyperparameters in Tab.~\ref{tab:ablation_all}.
For the Positive-Negative frame ratio, an 8:2 distribution achieves the best temporal and spatial balance.
An extreme ratio towards negative frames (2:8) decreases temporal grounding (dropping to 51.4) due to the lack of positive visual cues, while a fully positive sampling (10:0) lacks the necessary discriminative penalty. 
Regarding the loss weighting between the autoregressive token loss ($\lambda_1$) and the spatial grounding loss ($\lambda_2$), a 1:0.02 ratio proves optimal. Equal weighting (1:1) over-penalizes the fine-grained spatial loss gradients, hurting overall performance.

\section{Conclusion}
In this paper, we proposed Bridge-STG, an end-to-end decoupled MLLM framework to resolve entangled spatio-temporal alignment and dual-domain visual token redundancy in STVG.
To overcome the semantic gap caused by architectural decoupling, we introduced the Spatio-Temporal Semantic Bridging mechanism with Explicit Temporal Alignment to distill the temporal reasoning context of the MLLM into robust bridging queries.
Guided by these queries, our Query-Guided Spatial Localization module uses multi-layer interactive queries and Positive/Negative Frame Sampling to achieve precise spatial grounding.
Extensive experiments demonstrate that Bridge-STG achieves state-of-the-art performance on STVG benchmarks and exhibits remarkable cross-task performance across VTG, VOT, REC, and VQA tasks, providing an effective architecture for multiple fine-grained video comprehension tasks.

\begin{acks}
 This work was supported in part by Fundamental and Interdisciplinary Disciplines Breakthrough Plan of the Ministry of Education of China JYB2025XDXM103, Alibaba Innovation Research (AIR) Program, Tencent WeChat Research Program, and SJTU-Huawei Explore X Gift Fund.   
\end{acks}

\bibliographystyle{ACM-Reference-Format}
\balance
\bibliography{sample-base}

\end{document}